\documentclass[journal]{IEEEtran}

\usepackage[colorlinks]{hyperref}

\usepackage{times}
\usepackage{amsmath}
\usepackage{graphicx}
\usepackage{graphics}
\usepackage{blkarray}
\usepackage{booktabs}
\usepackage[pdftex,dvipsnames]{xcolor}  
\usepackage{bbold}
\usepackage{multirow}
\usepackage{comment}
\usepackage{epstopdf}
\usepackage{algorithm2e}
\usepackage{tikz}
\usetikzlibrary{arrows,matrix,positioning}
\usepackage{afterpage}
\usepackage[labelformat=simple]{subcaption}
\usepackage{soul}
\usepackage{MnSymbol}
\usepackage{todonotes}
\usepackage{wrapfig}
\usepackage{units}
\usepackage{paralist}

\graphicspath{{./Pictures/}}
\usepackage{xargs}                      

\newcommandx{\revise}{\color{black}}
\newcommandx{\minor}{\color{black}}

\begin{document}

\title{A New UGV Teleoperation Interface With Network Connectivity Awareness}

\author{Ramviyas Parasuraman$^{1,\*}$, \and Sergio Caccamo$^2$, \and Fredrik B\aa berg$^2$, \and Petter \"Ogren$^2$, \and Mark Neerincx$^3$

\thanks{$^1$ Department of Computer and Information Technology, Purdue University, West Lafayette, IN 47907, USA.

$^2$ Centre for Autonomous Systems, School of Computer Science and Communication, KTH - Royal Institute of Technology, Stockholm 10044, Sweden.

$^3$ TNO Netherlands.

$^*$ Corresponding author email: {\tt ramviyas@purdue.edu}.}

\thanks{Authors retain copyright and grant the JHRI right of first publication with the work simultaneously licensed under a Creative Commons Attribution License that allows others to share the work with an acknowledgement of the work’s authorship and initial publication in this journal.}
}
\markboth{Accepted for publication in the Journal of Human-Robot Interaction (JHRI), Dec 2017.}%
{Min \MakeLowercase{\textit{Min et al.}}: A Directional Antenna based Leader-Follower Relay System for End-to-End Robot Communications}

\maketitle
\begin{abstract}
A reliable wireless connection between the operator and the teleoperated Unmanned Ground Vehicle (UGV) is critical in many Urban Search and Rescue (USAR) missions.
Unfortunately, as was seen in e.g. the Fukushima disaster, the networks available in areas where USAR missions take place are often severely  limited in range and coverage. Therefore, during mission execution, the operator needs to keep track of not only the physical parts of the mission, such as navigating through an area or searching for victims, but also the variations in network connectivity across the environment.

In this paper, we propose and evaluate a new teleoperation User Interface (UI) that includes a way of estimating the Direction of Arrival (DoA) of the Radio Signal Strength (RSS) and integrating the DoA information in the interface. The evaluation shows that using the interface results in more objects found, and less aborted missions due to connectivity problems, as compared to a standard interface.

The proposed interface is an extension to an existing interface centered around the video stream captured by the UGV. But instead of just showing the network signal strength in terms of percent and a set of bars, the additional information of DoA is added in terms of a color bar surrounding the video feed. With this information, the operator knows what movement directions are safe, even when moving in regions close to the connectivity threshold.
\end{abstract}

\begin{IEEEkeywords}
Teleoperation, UGV, Search and Rescue, FLC, Network Connectivity, User Interface.
\end{IEEEkeywords}

\IEEEpeerreviewmaketitle 

\section{Introduction}
\label{sec:intro}

Today, teleoperated UGVs play an increasingly important role in a number of high risk applications, including 
USAR and Explosive Ordinance Disposal (EOD).
The successful completion of these  missions depend on a reliable communication link between operator and UGV, but unfortunately experiences from Fukushima and the World Trade Center disaster show that cables can limit performance, or break~\cite{Nagatani2013}, and wireless network connectivity can be lost~\cite{Murphy2014d}.

It is reasonable to believe that the very nature of USAR scenarios imply a high risk of damages to infrastructure, including electricity and wireless network facilities. To avoid relying on a stable network connection, one possible solution would be to enable the UGVs to operate autonomously, but for the foreseeable future,
human operators will remain more versatile than autonomous systems when it comes to decision making, in particular in challenging and unpredictable USAR environments \cite{Wegner2006,Muszynski2012a}. 

{\minor \cite{yanco2004acquiring} defines the Situation Awareness (SA) in the context of human-robot interaction as : {\it the perception of the robots' location, surroundings, and status; the comprehension of their meaning; and the projection of how the robot will behave in the near future}.
Thus, the connectivity awareness is viewed as a component of SA (network status), determining where the robot can operate. 
}

\begin{figure}[t]
  \center
     \includegraphics[width=0.95\columnwidth]{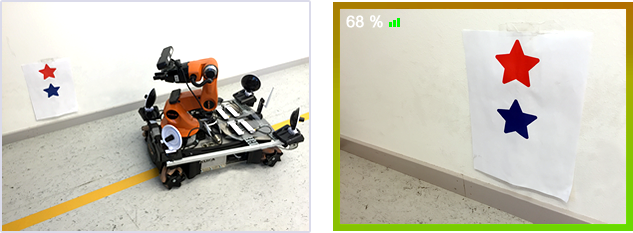}
  \caption{{ The youBot mobile robot equipped with wireless network hardware used in the experiments (left) shown along with the user interface (UI) displaying the RSS DoA as a color bar around the video feed from the robot.}}
  \label{fig:HMI_youbot}
\end{figure}

In this paper, we address the problem of improving SA such that the operator is aware of dynamic network connectivity status and adjust the UGV operation to it.
This is done by extending the user interface (UI) with not only a measure of Radio Signal Strength (RSS), but also a notion of the motion direction (i.e. the DoA) that would increase this signal strength, and thereby the communication quality (delay, packet loss, etc.) which is known to affect teleoperation task performance \cite{Rank2016,owen2013haptic}.

Using the proposed solution, an operator close to the connectivity limit knows which way to go to improve the connection. An operator who, for example, would like to move the UGV a bit more to the left to inspect a cavity, knows if this move will improve, worsen or leave the RSS unchanged. 

The proposed UI is composed of two parts, first the DoA is estimated, then it is presented to the operator in an 
efficient manner. 
The estimation of the DoA is done by using spatially dispersed wireless receivers on the four edges of the UGV (as can be seen in Fig.~\ref{fig:HMI_youbot}) and applying the finite differences method to extract the RSS gradient. We then employ spatial and temporal filtering schemes to mitigate  multipath fading effects and transient noises in the  measurements. The estimation and  filtering algorithms run online and dynamically adapt to changes in the wireless environment, such as a change in network connection (e.g. introduction of an intermediate relay robot as a signal repeater) or movement of a mobile wireless access point connecting the robot to the base station.

\begin{figure}[t]
	\center
	\includegraphics[width=0.95\columnwidth]{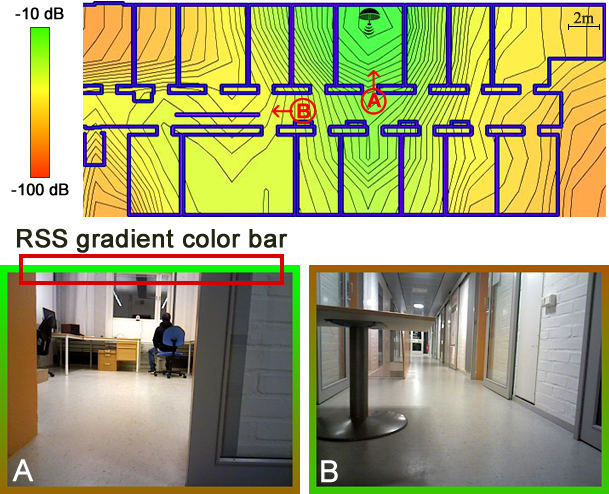}
	\caption{{
			A map of the RSS in an office environment and two examples of the UI with the UGV at positions A and B.
			Note the green and red gradient, indicating higher and lower signal directions (DoA) in the color bars surrounding the videos. 
	}}
	\label{fig:HMI_visualization}
\end{figure}

The presentation of the DoA to the operator was chosen in view of the fact that gaining a good SA is very challenging in USAR missions \cite{larochelle2012multi}. 
In fact, it was shown in \cite{burke2004mme,yanco2004acquiring} that as much as 49\% of mission time is normally devoted to improving the operator SA. 
Further, it was recommended in \cite{yanco2007rescuing} to use a large central part of the screen for the video feed.
Therefore, we propose to represent the DoA information in the form of a color bar surrounding the video feed (see Fig.~\ref{fig:HMI_visualization}) to provide SA to the operator in terms of connectivity status and physical surroundings.
{\minor Note that there are many possible variations on the proposed idea of graphically illustrating the DoA, including arrows of different forms and placements. However, we focus the investigation on the potential benefits of providing such information. Comparing different variations on the theme is beyond the scope of this study.}

For the evaluation, we  identified two important challenges associated with teleoperation of UGVs in USAR missions: (1) providing effective SA to the operator and (2) ensuring resilient wireless connectivity with the UGV. High SA can reduce mission time and improve operator decisions, while a resilient network connection will avoid losing control of the~UGV.

The main contributions of this paper are three-fold. We first propose a new way of estimating DoA for teleoperated UGVs. We then propose a way of integrating this DoA information in a UGV teleoperation UI. Lastly, we perform a user study, showing that the proposed approach in fact increases the number of found objects during a search mission, and decreases the chances of losing the connection to the UGV. To the best of our knowledge, none of these items have been done in a UGV teleoperation context before. 
This paper extends our previous work \cite{Caccamo2015}, with an improved design and a thorough evaluation of the proposed interface.

The paper is organized as follows. First, Section \ref{sec:relatedWork} reviews the literature on this topic and Section~\ref{sec:background} describes the proposed approach. Then, Section~\ref{sec:evaluation}
describes the human in the loop experiments, 
with results in Section~\ref{sec:results} and the discussion in Section~\ref{sec:discussions}.  Finally, we conclude in Section~\ref{sec:conclusions} and provide suggestions for further work. 

\section{Related Work}
\label{sec:relatedWork}

The wireless network connectivity of USAR UGVs have often proved unreliable \cite{murphy2004hri,Carlson2005}, with examples including real incidents where robots were lost during disaster inspection operations \cite{Nagatani2013,Murphy2014d}.
Casper et al. \cite{Casper2003} investigated user confidence in remotely operated robots with intermittent communications, and found that these problems had a significant impact on  the usability of the systems. They even suggested that because of communication dropout problems, wireless robots should be avoided. 
However, the flexibility of wireless systems compared to tethered robots still make them an important alternative in many applications. 

A natural way of avoiding loss of communications is to make the user aware of the connection quality.
 A decade ago, this information was usually not  displayed in the Operator Control Unit (OCU) \cite{Fong2001},
 but more recently, it is often added in the form of
 a "signal bar"  (as in modern cell phones) or in form of a percentage. Typical examples of such representation can be seen in \cite{Larochelle:2011wl,hedstrom2006} including the recent Quince 2 robot's OCU \cite{yoshida2014}. Furthermore, the Wayfarer OCU for Packbot robots \cite{yamauchi2004packbot} represent the radio signal level in a vertical bar manner, in addition to a numeric indicator.

The literature on robot interfaces also include examples where information about gradients and directions is 
made available to the user.
In \cite{Hestand2004,baker2004improved}  two microphones on the left and right of the robot were used to estimate the direction of a sound source, which was displayed (overlaid on the video) in the form of a pointer floating on horizontal and vertical lines. A similar representation was used in \cite{hedstrom2006} to show robot speed information. In \cite{de2011enhancing}, the authors proposed a tactile belt that vibrates in the direction of detected collisions to improve SA, while in \cite{Smets2008} a study found that the use of a tactile vest did not improve SA significantly in navigation tasks.

An influential study in  Human-Robot Interface (HRI) design \cite{yanco2007rescuing} advocates the use of a large single interface with a significant percentage of the screen dedicated to video. The authors also recommend  providing more spatial information about the environment to increase SA, and using fused sensor information to lower the cognitive load on user. Moreover,  multi-sensory interfaces had also been advocated in the literature \cite{de2014multi}.

In this paper we go beyond the related work described above by having the teleoperation interface include not only a scalar value to describe the network connectivity situation, but also the direction in which it is expected to improve, i.e. the DoA. Assessing the geographic distribution of network connectivity is a spatial task, for which the visual modality fits best with the human information processing, see e.g.~the multi-resource model of Wickens \cite{Wickens2008}. Therefore we choose to present the DoA in the  form of visual gradient bars surrounding the video feedback.  

Carefully integrating the DoA information into the visual feedback is crucial. For this we use  FLC (Free Look Control)  \cite{Ogren:2013kx}  as the control layer. FLC is essentially a "navigate-by-camera" mode as envisioned in \cite{yanco2006analysis}. In the FLC mode, the operator controls the UGV in relation to the camera frame instead of the world frame, making it more intuitive than the traditional so-called \emph{Tank Control} mode. Hence it is appropriate to use FLC for presenting the DoA information in direct reference to the camera frame, making the UGV control easier while simultaneously enhancing local SA. 
\section{The proposed approach}
\label{sec:background}
\label{sec:approach}

The new user interface, as most robot teleoperation UIs, is composed of two parts: receiving control commands from the operator; and providing  feedback to the operator. In the former, we use a gamepad controller and  the FLC interface \cite{Ogren:2013kx} and in the latter we present the DoA as an extra sensory feedback in addition to the video stream. The hardware configuration and associated signal processing required to realize the new interface 
is presented below. 

\subsection{FLC interface}
\label{sec:FLC}

FLC, a UGV control interface inspired by the First Person Shooter (FPS) video games genre, combines  camera and platform control, thus permitting the operator to completely focus on commands for moving the UGV camera (the UGV adjusts its heading accordingly) through the remote environment (world frame) without worrying about the orientation of the UGV chassis. This is in contrast with the standard interface, \emph{Tank Control}  which is used in most of the teleoperated UGVs today, where the operator is required to mentally keep track of at least two orientations while teleoperating an UGV: the camera angle relative to the UGV, and the platform orientation with respect to the world frame. 
The advantages of FLC compared to Tank Control were investigated in \cite{Baberg2016},
and  more details about implementing FLC can be found in  \cite{Ogren:2013kx}.

\subsection{DoA estimation}
\label{sec:DoA}
\begin{figure}[t]
	\center
	\includegraphics[width=4.5cm]{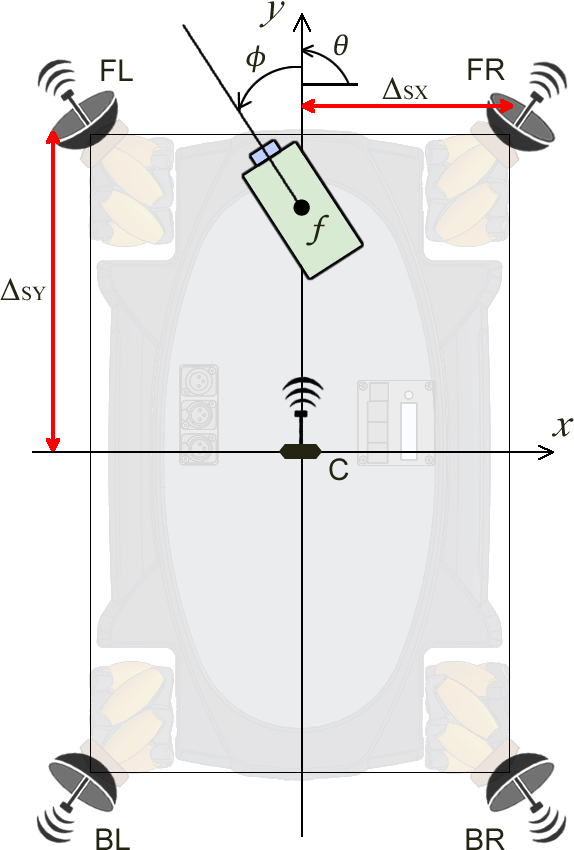}
	\caption{{UGV equipped with a camera, one wireless adapter at the center and four wireless adapters with directional antennas at the corners.}}
	\label{fig:ugv}
\end{figure}
A good estimate of the DoA forms a core part of the new interface. 
It has been shown that the DoA can be estimated by the direction of the  RSS gradients  \cite{Han2009}. For calculating RSS gradients, we use four wireless adapters connected to external directional antennas placed on the corners of the UGV\footnote{The squared planar arrangement of antennas is suggested in \cite{Parasuraman2013} due to its robust nature. Moreover, the directional antennas are chosen due to high stability and accuracy in the measurement and higher link throughput \cite{Min2013}.} as shown in Fig.~\ref{fig:HMI_youbot} and \ref{fig:ugv}. We also use a fifth wireless adapter connected to an omnidirectional antenna. The former four adapters are used for DoA estimation whereas the latter is used for actual communication for the teleoperation between the UGV and the control station. 

We measure the RSS (in dBm) from all the wireless adapters using the Received Signal Strength Indicator (RSSI\footnote{RSSI is a vendor-specific metric and therefore reports different values (or quantities) in different devices. The wireless adapters used in this paper reported reliable values of absolute signal power (dBm) as RSSI.}) metric which is usually prone to noise and temporal variations (due to environmental dynamics) \cite{Lindhe2007,Rappaport2001}. These noisy fluctuations are mitigated by applying an exponentially weighted moving average (EWMA) filter on the measured RSS from each wireless adapter using the following model \cite{Parasuraman2014b}: 
\begin{equation}
{R^f(i)} = R^f(i-1) + \alpha (R(i)- R^f(i-1)),
\end{equation}
where $R(i)$ is the RSS value measured at the $i^{th}$ instant, $R^f$ is the filtered RSS value and $\alpha$ is an empirical smoothing parameter. In addition to the EWMA filter, a Moving Average Filter (MAF) is also applied to mitigate spatial multipath fading, with a window size equal to about $10{\lambda}$ ($\lambda$ being the wavelength) as suggested in \cite{Valenzuela1997}. The MAF window depend on the UGV velocity, RSS sampling frequency and the wavelength of the radio signal\footnote{For instance, if the UGV velocity is $\unitfrac[0.2]{m}{s}$, RSS sampling frequency is $\unit[5]{Hz}$, and using $\unit[2.4]{GHz}$ signal (wavelength $\lambda = \unit[12.5]{cm}$), the MAF window size should be $\approx 30$ to filter samples within \unit[1.25]{m} ($10\lambda$) displacement of the UGV.}.

Modeling the RSS as a two-dimensional scalar field\footnote{
Being a ground vehicle, the UGV locally moves in a plane.} 
it is possible to obtain the gradient of the RSS field ($\overrightarrow{g} = [g_x,g_y]$) with respect to the center of the UGV using the central finite difference method \cite{parasuraman2013spatial,Bezzo2014}:

\begin{equation}
\begin{aligned}
g_x =  \frac{(R_{FR} - R_{FL})}{ 2\Delta_{SX}} + \frac{(R_{BR} - R_{BL})}{2\Delta_{SX}} , \\
g_y =  \frac{(R_{FR} - R_{BR})}{2\Delta_{SY}} + \frac{(R_{FL} - R_{BL})}{2\Delta_{SY}} , \label{eq:RSSgrad}
\end{aligned}
\end{equation}

where $\Delta_{SX}$, $\Delta_{SY}$ are the corresponding spatial separations between the antennas, $R_{FR}, R_{FL}, R_{BR}$ and $R_{BL}$ are the filtered RSS values of the Front-Right (FR), Front-Left (FL), Bottom-Right (BR), Bottom-Left (w.r.t the center of the UGV) receivers respectively, as can be seen in Fig.~\ref{fig:ugv}. The orientation of each antenna is aligned with its placement. It is possible to employ redundant gradient estimation methods to tackle device failures or misreadings, as discussed in \cite{Parasuraman2014b}.

From the RSS gradient ($\overrightarrow{g}$), the DoA of the radio signal is obtained as,
\begin{equation}
\text{DoA}_{\theta} = \tan^{-1} ( \frac{g_{y}}{g_{x}} ).
\label{eqn:DoA}
\end{equation}

\subsection{User Interface}

A large video feed on the OCU is used for  teleoperating the UGV. 
This visual interface is extended to include wireless connectivity information by adding a colored gradient bar surrounding the real time video feed from the UGV. The added color bar illustrates the DoA relative to the camera view.
This setup was inspired by computer game interfaces, where the direction of threats causing health level changes is communicated using
  flashing colors in the appropriate part of the screen. Consequently, the new interface can be categorized as Type 3.2.2.1 (additional visual input: type - communication level) in the framework for analyzing human robot interaction \cite{richer2006video}. 

In the UI, we create a rectangular border around the video, as illustrated in Fig \ref{fig:HMI_visualization}. As the DoA computed with Equation~\eqref{eqn:DoA} is first given in the UGV frame, it is converted to the camera frame (to provide a first person view of the DoA to the operator){\minor .}
Then we translate the DoA to a color gradient bar around the camera view by scaling the color intensity according to a linear interpolation of the measured RSS values around the corners. A green color in the color bar  indicates the higher signal strength direction, whereas a red color indicates a lower signal strength direction. Thus the interface not only represents DoA but also gives a sense of the absolute RSS. 

{\revise 
\subsection{Experimental verification}
\label{sec:Experiments}

In \cite{Caccamo2015}, we investigated the accuracy  of the proposed DoA estimate. Specifically, we performed experiments to verify that the variation of the RSS along a robot path is indeed predicted by the DoA estimates. We summarize the key findings in this section. 

Firstly, it was found that the DoA estimation had high accuracy in both Line-of-Sight and Non-Line-of-Sight conditions, the absolute mean DoA error was within 0.2 rad (12 degrees), an accuracy that will turn out to be good enough for our purposes. 

Secondly, a set of experiments were performed to evaluate the sensitivity, specificity, accuracy and precision of the DoA feedback provided by the interface. To gather data, the robot was teleoperated  by a human operator, simulating short missions following different paths. Eight different trials of this kind were conducted. During each trial, we logged the robot position data obtained from the dead-reckoning of the wheel odometers, the RSS data, the estimated DoA and the streamed video. The dead-reckoning of the wheel odometers was not very accurate, but this was not a problem as both motion directions and estimated DoA are given in the same local coordinate system. A video illustrating the proposed method with an example trial is available online\footnote{\url{https://youtu.be/YcbPi1c7eaQ}}. 
It can be seen in the video illustration that the estimated DoA sometimes pointed towards the corridor or the doorways (instead of the true source location). This is expected because of substantial exposure of radio signals from these regions.

In a noise free world the following equality would hold:
\begin{equation}
\frac{d R}{dt}=\frac{d R}{dx}\frac{d x}{dt} \;,
\label{eq:noise_free}
\end{equation}
where $x\in\mathbb{R}^2$ is the spatial dimension. The real world is however far from noise free, and we had to experimentally verify that our estimates provide useful information to the human operator.
For the estimates to be useful, the measured RSS should increase when the UGV is moved in the direction of the DoA, i.e. the two sides of Equation (\ref{eq:noise_free}) should have the same sign.
 We used temporal differences in the measured RSS at the central receiver ($R_C$) to estimate $\frac{d R}{dt}$, and $\overrightarrow{g}$ as the estimate of $\frac{d R}{dx}$ and the odometer robot velocity $\overrightarrow{\nu}$  to estimate $\frac{d x}{dt}$. 
The scalar (dot) product between the robot velocity and the computed RSS gradient at each instant is given by:
\begin{align}
p(t) = \langle \overrightarrow{g}(t) ,\overrightarrow{\nu}(t) \rangle  .
\end{align}

By comparing the scalar product $p(t)$ with the change in the RSS at the central receiver $\nabla_t R_C =  \frac{d R_C}{dt}$, we  evaluated the efficacy of the proposed  system. We expected a steep increase in $R_C$ when $p(t)$ is positive and close to 1 (i.e. when the user is moving towards the DoA). Similarly, we expected a sharp decrease in $R_c$ when the p(t) is negative and close to -1 (i.e. when the user moves the robot away from the DoA).
\begin{figure}[t]
	\center
	\includegraphics[width=0.95\columnwidth]{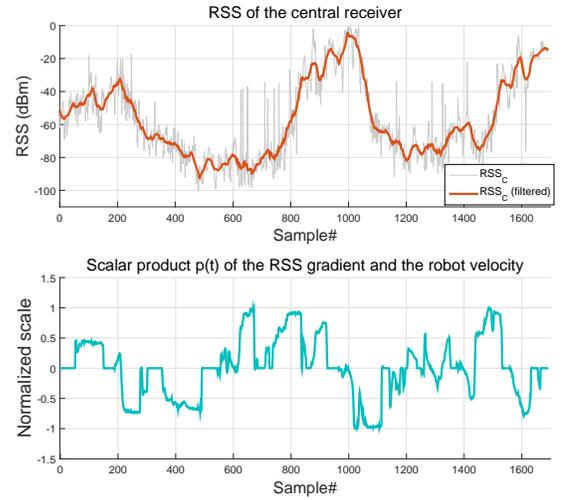}
	\caption{{Quantitative evaluation of the new UGV teleoperation interface with the RSS DoA feedback.
			The estimate is useful when the scalar product $p(t)$ has the same sign as the derivative of the RSS. i.e. the changes in the RSS values follow the directions indicated by $p(t)$.
	}}
	\label{fig:Robots-trajectory-scalar-product}
\end{figure}

Fig.~\ref{fig:Robots-trajectory-scalar-product} shows the variations of RSS at the central receiver (\(R_C\)) and the scalar product $p(t)$ with time for a sample trial. 
To quantify the system performance, we measured the number of true/false (T/F) positives/negatives (P/N) in the outcome. Using these measures, we computed Sensitivity ($\frac{TP}{TP + FN}$), Specificity ($\frac{TN}{FP + TN}$), Precision ($\frac{TP}{TP + FP}$), and Accuracy ($\frac{TP + TN}{TP + TN + FP + FN}$) metrics. 
 \begin{table}[h]
	\caption{Evaluation of the DoA feedback for sensitivity, specificity, precision and accuracy.}
        \centering
	\begin{tabular}{| c || c |c |c |c | }
		\hline
		\textbf{} & {Sensitivity} & {Specificity} & {Precision} & {Accuracy}\\ \hline
		{Mean} & 0.74 & 0.83 & 0.82 & 0.78 \\
		\hline
		\multicolumn{5}{c}{}  
	\end{tabular}
	\label{tab:ioq}
\end{table}

In Table~\ref{tab:ioq}, we present the key results obtained from the eight experiments with an average mission time of 9.2 minutes each. The proposed system delivered high accuracy and precision in guiding the teleoperator with network connectivity feedback in an indoor environment. As the analysis depended on the UGV's velocity from the odometer, we note that odometry errors could have impacted the analysis of the accuracy of the proposed system. Thus a better localization technique would have improved the overall system analysis. Note that the system has shown reasonable sensitivity in directing the operator into high wireless signal regions (towards DoA) while maintains high specificity in pointing out low-wireless signal regions.

Although the above quantitative results are fairly promising, a qualitative evaluation with user studies is required to investigate the effectiveness of the overall system. This will be done in the following sections.
}
\section{User evaluation}
\label{sec:evaluation}

To evaluate the actual system performance of the new interface, we conducted experiments with human subjects. The experimental setup consisted of an exploration task (search for symbols) with a remotely operated UGV platform in an unfamiliar maze-like environment. The objective of the experiments was to evaluate the new visual DoA interface (denoted  \emph{VDOA} hereafter) against the (state-of-the-art) standard OCU interface that displays radio connectivity using a signal bar and percentage value (denoted \emph{BAR}). The two interfaces are shown in Fig.~\ref{fig:interfaces}. To allow a fair comparison, both interfaces used FLC as control layer. 
{\revise Note that we are interested in the evaluation at the first two levels of SA (perception, comprehension) \cite{endsley2000} {\minor because the proposed DoA interface does not predict the future connectivity status (Level 3 - prediction). Nevertheless, {\it VDOA} allows the operator to infer the present and future network availability in different travel directions.}

\subsection{Evaluation framework}
\label{sec:evaluationintro}

When designing the experiments, we followed the situated cognitive engineering (sCE) method \cite{sce} in which we first identified the two \emph{core functions} that we want to compare, see items 1 and 2 below. 
Then we formulated a number of \emph{claims}, i.e. hypothesis connected to the \emph{core functions},
listing a number of possible \emph{upsides} (benefits) and \emph{downsides} (drawbacks)  of each hypothesis.
These up/downsides are then used to define \emph{what to measure} in the experiments.
Finally, having performed the experiments we can then see which of the possible \emph{up/downsides} are
confirmed by data, and hence draw conclusions about the \emph{claims} and when the different
  \emph{core functions} can be beneficial.

The following core functions describe what the corresponding systems do. 
\begin{enumerate}
 \item \emph{VDOA provides a graphical indication of the DoA and  the  RSS value  in the periphery of the teleoperation display. See Fig.~\ref{fig:interfaces} (left). It also shows the RSS value in the same way as BAR below.}
 \item
 \emph{BAR shows the RSS value in the form of both a percentage and a set of signal bars.  See Fig.~\ref{fig:interfaces} (right).}
\end{enumerate}

\begin{figure}[t]
\center
\includegraphics[width=0.95\columnwidth]{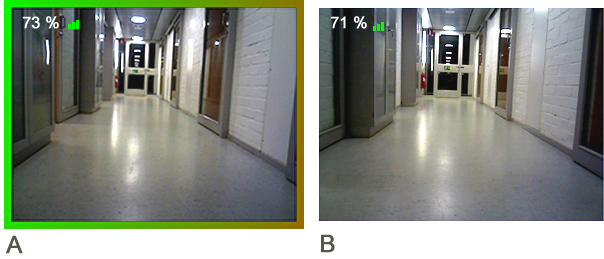}
\caption{{Visualization of the proposed VDOA (A) and the standard BAR (B) interfaces to represent connectivity information.}}
\label{fig:interfaces}
\end{figure}
Listed below are the claims that we make on the core functions, with corresponding upsides/downsides (U/D) and what to measure in parenthesis.

\begin{enumerate}[\bfseries {Claim} 1{:}]
\item VDOA leads to UGV trajectories in  higher signal strength regions.
\begin{itemize}
	\item U11: Less error in the estimated DoA ({\it radio source localization})
	\item U12: Increased connectivity and less connection loss ({\it signal strength, loss of connectivity})
	\item U13: More useful area covered and more time spent during exploration {\minor due to less connection loss}. ({\it coverage, execution time})
	\item D11: Less concentration on {\minor the surroundings such as objects and obstacles in the robot proximity e.g. while the user follows the VDOA guidance to change direction} ({\it collisions, ease of finding symbols})
\end{itemize}
\item VDOA provides better situation awareness.
\begin {itemize}	
	\item U21: Better SA {\minor on the search task and connectivity status} ({\it situation awareness})
	\item U22: More symbols found ({\it symbols found}) 
 	\item U23: More spatial understanding ({\it symbols mapping accuracy, radio source localization}).
	\item D21: Higher mental effort {\minor due to additional information} ({\it mental effort}\footnote{Rating Scale of Mental Effort (RSME) \cite{zijlstra1985}})
\end {itemize}
\item VDOA improves user experience.
\begin {itemize}
	\item U31: Better usability and satisfaction {\minor of the useful DoA information on the interface} ({\it usability, preference}). 
	\item U32: Higher time utilization or longer time spent on actual tasks {\minor due to higher connectivity awareness and less connection loss} ({\it execution time, symbols found}).
	\item U33: Better understanding of the network connectivity across various regions ({\it radio source localization, loss of connectivity}).
    \end {itemize}
\item BAR increases focus and concentration on the actual task 
\begin {itemize}
	\item U41: Less collisions during exploration ({\it collisions})
	\item U42: Lower mental effort ({\it mental effort})
	\item U43: Easier to operate the robot and use the interface {\minor due to its simplicity} ({\it preference, usability})
	\item D41: More connection loss {\minor since it is more difficult to understand spatial (2D) wireless connectivity using BAR} ({\it loss of connectivity, signal strength})
	\item D42: Less situation awareness and less spatial understanding in terms of network status ({\it situation awareness, coverage, execution time, symbols mapping accuracy})

\end {itemize}
\end{enumerate}

These claims are used to argue which interface is better suited for UGV teleoperation, especially in a USAR scenario. The upside or downside of a claim can be confirmed if it is supported by at least one of its measures. The validation of these claims help us to determine the interface that is effective (maintains better connectivity), productive (higher task utilization, higher coverage areas), and is more appreciated by the operators (preference and better usability).

\subsection{Method} 
\subsubsection{Experimental design}
Considering that the radio propagation in a given environment is unique to specific settings, we conducted \emph{"between-subjects"} trials instead of \emph{"within-subjects"} trials for comparing the two interfaces. This is due to the very high probability of carryover effects associated with the memory of radio signal coverage if a \emph{"within-subjects"} design is performed. Therefore, following a \emph{"between-subjects"} design, $N$ participants in each interface group (VDOA and BAR) are recruited for executing tasks based on a set of instructions (explained below). The nature of the opponent group/experiment is revealed to the participants only at the end of the experiment (and survey) to avoid biasing effects.
 
\subsubsection{Participants}
Based on statistical expectations on the outcome and the characteristics of the measured variables, the results of sample size and power calculations reveal that at least eight\footnote{This number was derived using the standard power tests \cite{Dell2002} assuming a power level of 80\% and false positive rate of 5\% with at least 20\% difference in the expected means of the two groups (with a standard deviation of 20\%).} participants is required for each group. Thus a total of at least 16 participants were required. However, we recruited a total of 24 participants for this study to increase the power. The participants (15 male and 9 female) were all university students and staff in the same age group (mean age: 27.9). Most of the participants did not have prior experience with robots or UGVs (mean experience: 2.04 out of 5). Although we  conducted the user study with 24 participants, the data of 4 participants were not useful because of technical issues such as motor drive fault, operator fault, etc. faced during the experiment. Therefore, we used the data of 20 participants (12 male, 8 female) with ten ($N=10$) in each control group in our analysis.

\subsection{Variables}
\label{sec:variables}
In accordance with the claims to be tested, Table~\ref{tab:variables} lists the variables measured in the experiments.
In the \emph{How} column, the way of collecting the measurements is indicated as data logging in the real robot (Datalog), through manual observations (Obs), or through a questionnaire (Q). The \emph{Claim} column shows the associated claims (upsides/downsides) of each measurement. 
\begin{table}[t]
\caption{Measurement variables used in the user evaluation.}
\begin{center}
	\begin{scriptsize}
		\begin{tabular}{|l | l | l |}
			\hline
			\bf{Measurement} & \bf{How?} & \bf{Claim} \\
			\hline
			Subjective & & \\
			\hline
			Reported overall usability   & Q  & U31, U43  \\
			Mental effort   &Q  &  D21, U42 \\
			Preference 	&Q	&	U31,U43 \\
			Ease of finding symbols &Q & D11 \\
			Situation Awareness (exploration, network)   & Q &  U21, D42  \\
			\hline
			Objective & & \\
			\hline
			Number of symbols found  &Obs & U22, U32, D42  \\
			Situation Awareness (spatial - symbols mapping) &Q+Datalog & U23, D42  \\
			Number of collisions   & Obs   &  D11, U41  \\
			Execution time &Obs+Datalog & U13, U32  \\
			Localization of radio source &Q+Obs & U11, U23, U33  \\
			Coverage (area/distance)  &Datalog & U13, D42  \\
			Number of connection losses & Datalog  &  U12, U33, D41  \\
			Radio Signal Strength (RSS) & Datalog  &  U12, D41  \\
			\hline 
			\multicolumn{3}{c}{} 
		\end{tabular}
	\end{scriptsize}
\end{center}
\label{tab:variables}
\end{table}

\subsection{Test environment}
\label{sec:testenv}
\subsubsection{Procedure}
\label{sec:procedure}
Written and verbal instructions were given to the participant at the beginning. Participants then had to answer (fill in) general questions on their experiences with robots and games. Then they were informed about the experiment as per the instructions. This was followed by a training session where the users were asked to drive the UGV in a rectangular path in a small room without colliding. The user was also  given the real position of the radio transmitter (used for training)  in order to assess the connectivity information in the UI. The training session lasted until the users expressed comfort in using both the FLC for control and the UI for perception\footnote{All training sessions lasted between 2-5 min.}. The real evaluation experiments commenced after the training session. The evaluation \emph{task} is explained below. Note that we used two wireless routers placed in different positions, one for training and the other for the actual task. 

{\revise After completing the experiments, the participants were asked to complete questionnaires on their experiences, situation awareness\footnote{\revise The key questions related to SA (on a scale of 1(No/Hard) to 5(Yes/Easy)) are the following: I have found all the symbols; I had enough time for exploring the area; How difficult was it to find the objects in the environment?; I think I have drawn the positions of the source correctly into the map; I think I have drawn the end position and orientation correctly; How difficult was it to find the source in the environment?}, metal effort, and various other factors that are listed in Table \ref{tab:variables}  with the label "Q". The participants were also asked to indicate on a map similar to the one in Fig.~\ref{fig:maze}, the location of symbols they found, the estimated radio source location, and the path taken by the UGV including the end position and orientation. }

\subsubsection{Hardware}
We used the same hardware and experimental setup as in \cite{Caccamo2015}.
\begin{figure}[t]
\center
\includegraphics[height=8.5cm]{{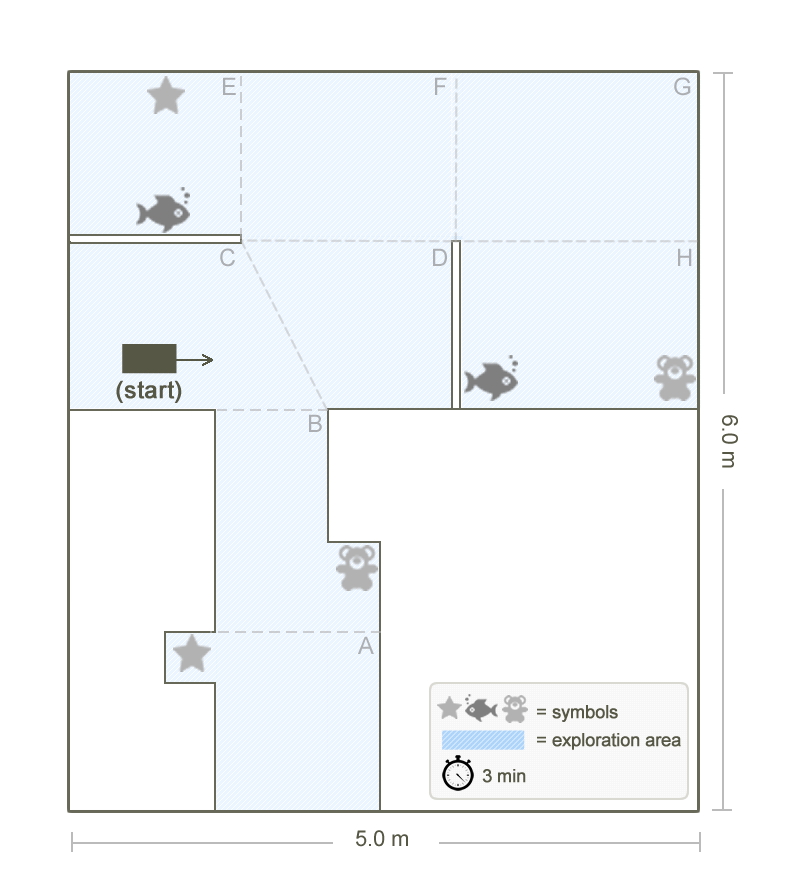}}
\includegraphics[height=8cm]{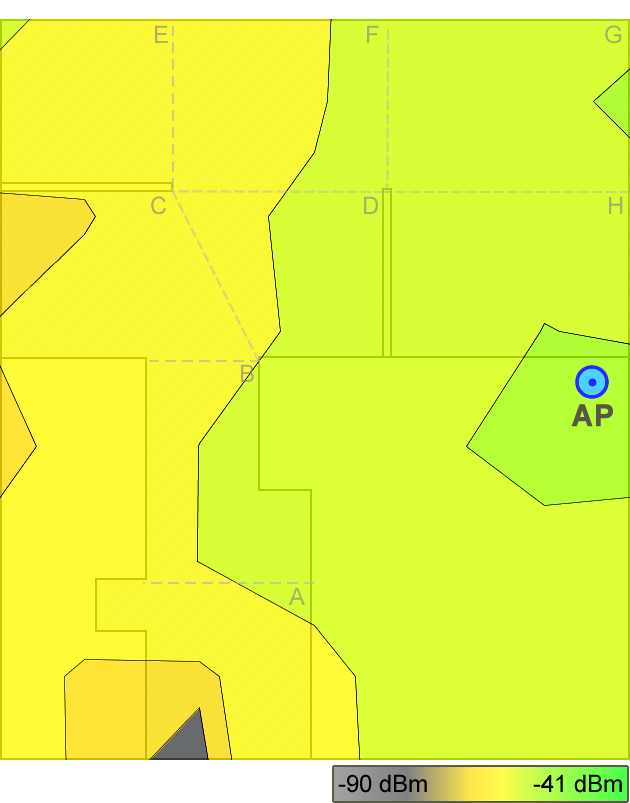}
\caption{{(Top) The map of the task scenario where the participant is asked to find symbols by exploring a given maze area. (Bottom) Snapshot of RSS map captured using a commercial wireless site survey tool from Ekahau. The radio source (WiFi router) location is marked as AP. }}
\label{fig:maze}
\end{figure}
\subsection{Task}
\label{sec:task}

The participants were asked to drive around an indoor environment (as they are more challenging for wireless signals) to search for known symbols as depicted in Fig.~\ref{fig:maze}. For this, a specially built maze was used. The maze is virtually split into eight regions as indicated with dotted lines. A time limit (3 minutes) was given to find symbols within the maze. The symbols shown in the figure have an area of approximately $\unit[40]{cm^2}$ and are placed on the walls of the maze with full visibility when the camera is aimed at them. 

The goal of the operator is to find as many symbols as possible without losing connectivity to the UGV. 
The experiment was stopped when either
 a timeout period was reached, or when the user had lost connectivity. The participant has no direct line of sight with the UGV and the only source of SA is the UI. 

During the task, data such as odometry, RSS, and loss of connectivity were recorded in a datalog. Execution time and the number of collisions were observed by a supervisor. 
The actual location of the radio source is indicated as AP on the map in Fig.~\ref{fig:maze} (which was not revealed to the participants). Often, the regions A and B experienced poor connectivity with high probability to lose connection,  whereas the regions C and E experienced average connectivity but had a lower probability to lose connection. Finally  D, F, G and H experienced high connectivity levels. Thus, as will be seen, how and when the regions A and B are approached turned out to be crucial to mission performance.
{\revise Note that, as can be seen in Fig.~\ref{fig:maze}, the symbols are placed in a manner such that they are equally distributed in different connectivity zones (poor, medium, high) of the exploration area. 
}
\section{Results}
\label{sec:results}

Fig.~\ref{fig:evluationresults} presents a boxplot result of the important variables. A summary of the user evaluation results can be found in Table~\ref{tab:resultTable}.  Below we describe the results in more detail, first in general, then specifically for the exploration task and finally the results related to the wireless network.

\begin{figure*}[ht]
	\centering
	\begin{subfigure}{0.24\textwidth}
		\includegraphics[width=\textwidth]{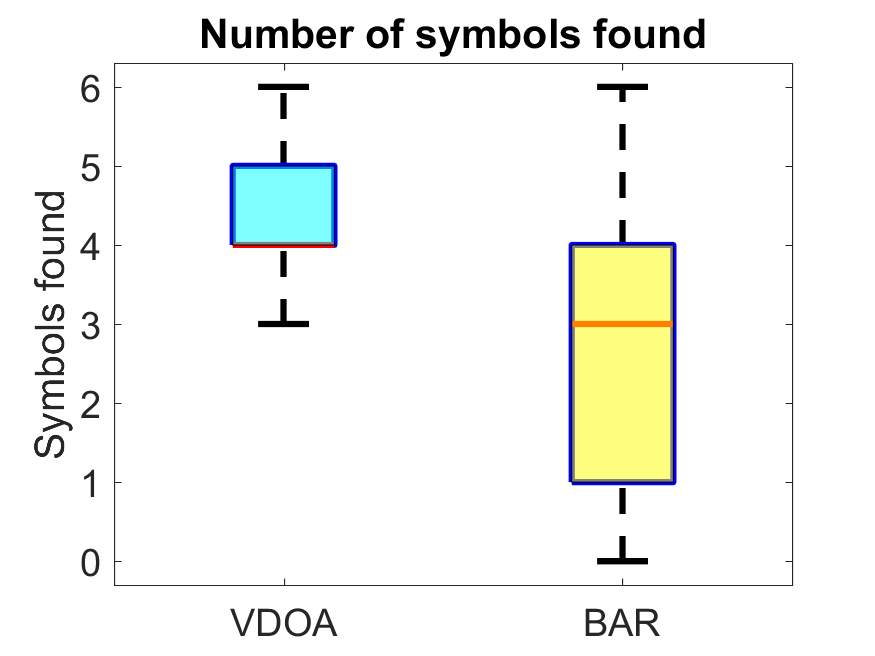}
	\end{subfigure}
	\begin{subfigure}{0.24\textwidth}
		\includegraphics[width=\textwidth]{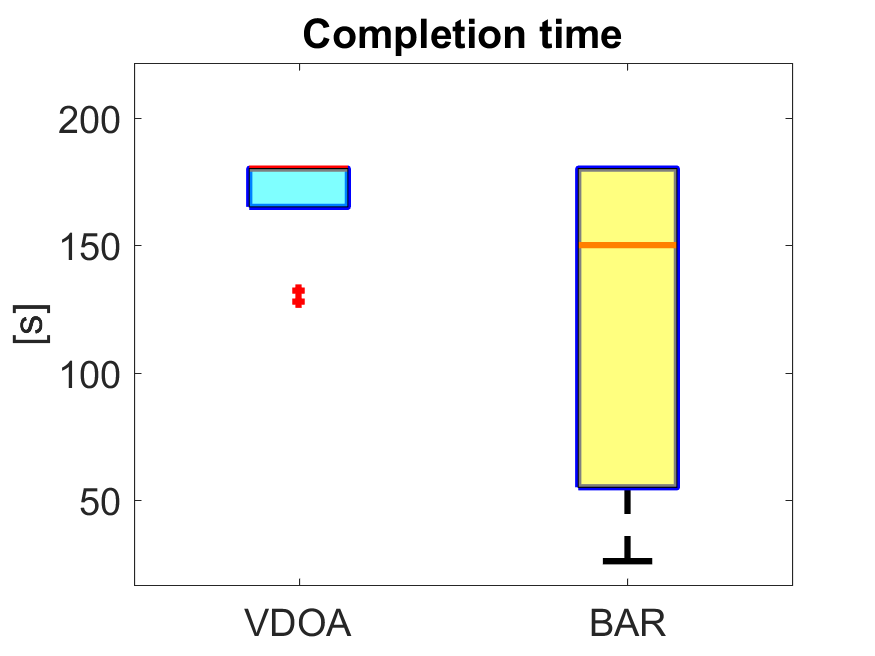}
	\end{subfigure}
	\begin{subfigure}{0.24\textwidth}
		\includegraphics[width=\textwidth]{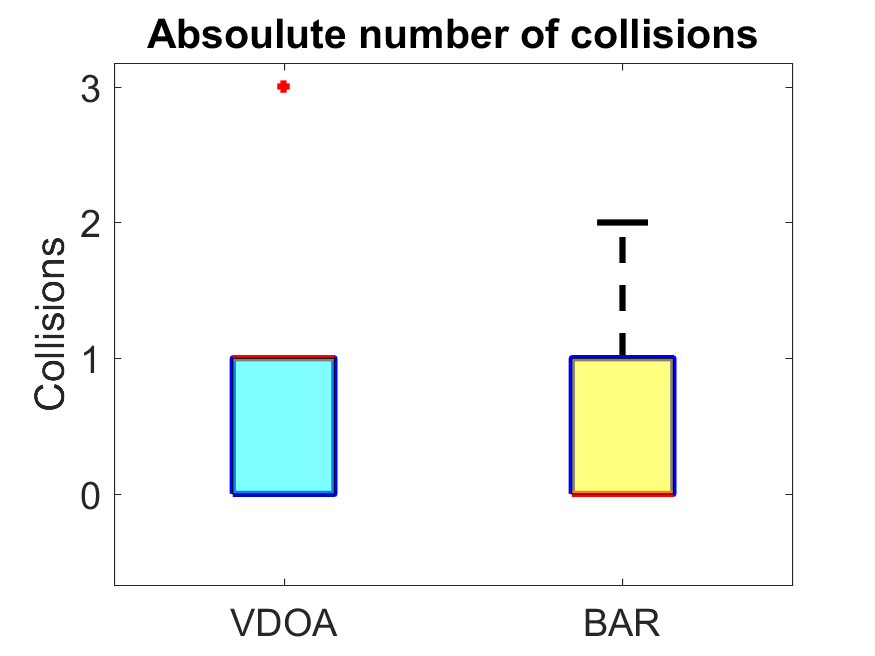}
	\end{subfigure}
	\begin{subfigure}{0.24\textwidth}
		\includegraphics[width=\textwidth]{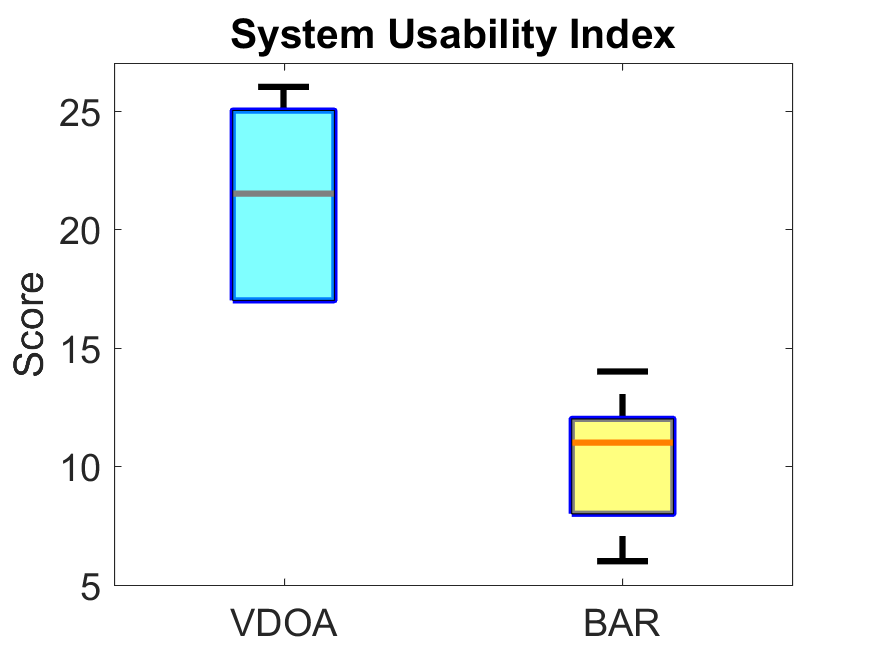}
	\end{subfigure}
	\begin{subfigure}{0.24\textwidth}
		\includegraphics[width=\textwidth]{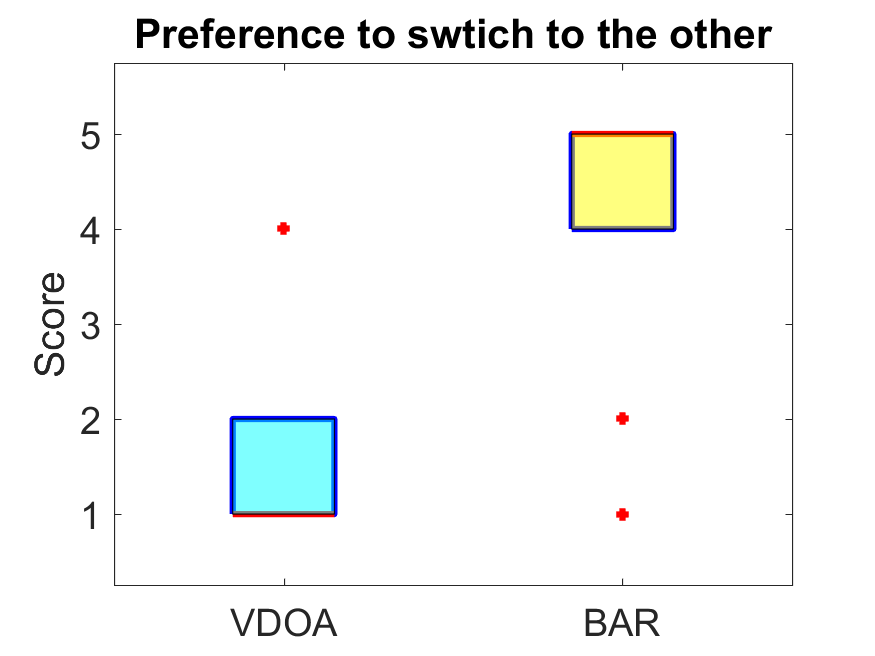}
	\end{subfigure}
	\begin{subfigure}{0.24\textwidth}
		\includegraphics[width=\textwidth]{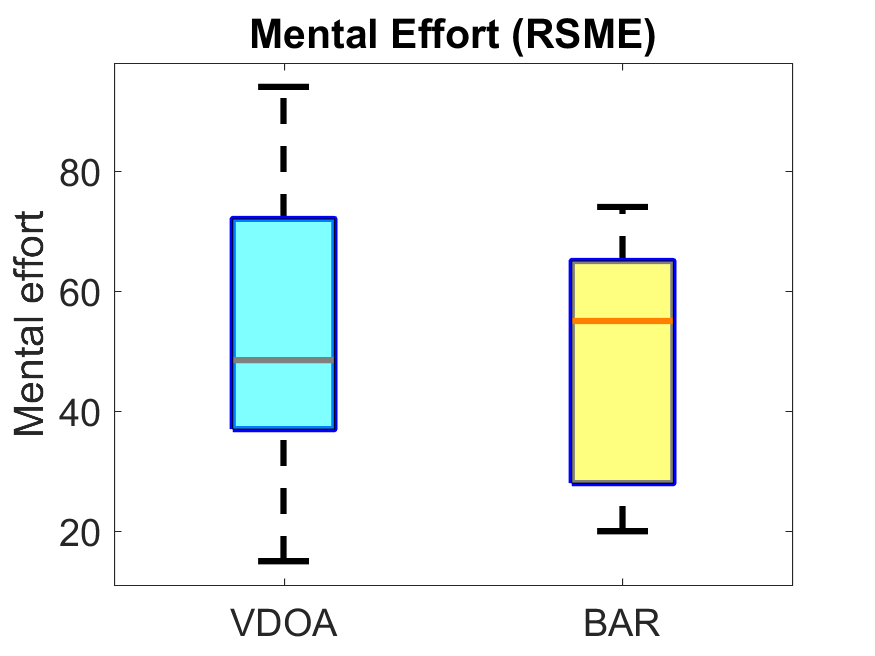}
	\end{subfigure}
	\begin{subfigure}{0.24\textwidth}
		\includegraphics[width=\textwidth]{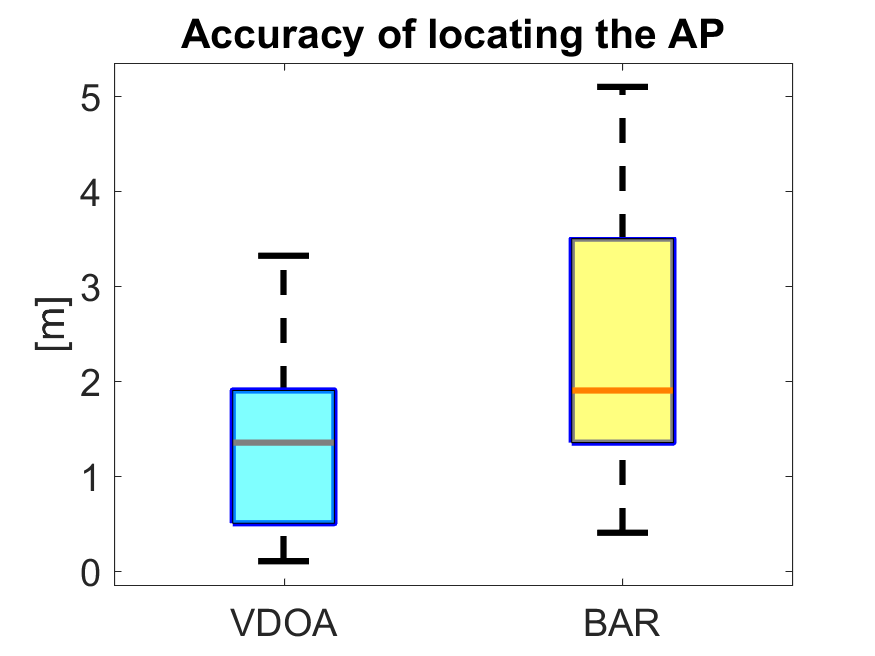}
	\end{subfigure}
	\begin{subfigure}{0.24\textwidth}
		\includegraphics[width=\textwidth]{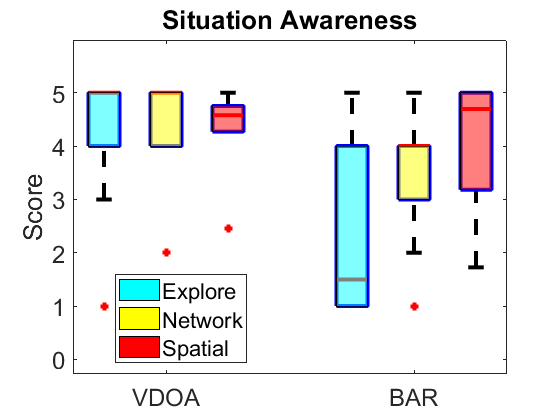}
	\end{subfigure}
	\caption{Resultant scores of important variables shown as boxplots. The red lines indicate the median values. {\minor A higher value indicates better {\minor results}, except in the following variables: {\it absolute number of collisions; preference to switch to the other; accuracy of locating the AP.}} }
	\label{fig:evluationresults}
\end{figure*}


\begin{table*}[ht]
\centering
	\caption{Summary of the user evaluation results. A (+) sign indicates relatively better value.}
	\begin{scriptsize}
		\centering
		\begin{tabular}{|l|c|c|c|c|c|c|c|c|c|}
			\hline
			\bf{Measurement} & \bf{Predicted} & \bf{Best in Eval.}&VDOA M& VDOA SD& BAR M& BAR SD & t(18) & \textit{p} \cr
			\hline
			\multicolumn{9}{|l|}{\bf{General measures}} \\
			\hline
			Reported usability    &-  &VDOA  & 21.3(+) & 3.88 & 10.4 & 2.91 &7.09  & $<$0.01 \cr
			Mental effort  &BAR  &No sign. res.  & 54.9 & 26.25 & 49.1(+) & 20.2 & -  & - \cr
			Preference    &-  &VDOA  & 1.8(+) & 1.23 & 4.1 & 1.45 & -3.82  & $<$0.01 \cr
			\hline
			\multicolumn{9}{|l|}{\bf{Task-related measures}} \\
			\hline
			No. of Symbols Found  &VDOA &VDOA  & 4.5(+) &0.97  & 2.6 & 2.07& 2.63 & $<$0.05 \cr
			Ease of finding symbols  &BAR &No sign. res.  & 4(+) &1.25  & 3.3 & 1.42 & - & - \cr
			Execution Time (s) & VDOA &VDOA    & 167.8(+) & 20.53 & 120.5 & 67.2 & 2.13 & $<$0.05 \cr
            No. of collisions & BAR &No sign. res.   & 0.8 & 0.92 & 0.4(+) &0.7 & - & - \cr
			Coverage (m) &VDOA &VDOA   & 8.78(+)  &2.84  & 5.28 &3.3 & 2.55 &$<$0.05 \cr
			Situation awareness (explore)  &VDOA &VDOA   & 4.2(+)  &1.32  & 2.5 &1.78 & 2.42 &$<$0.05 \cr
			{\revise Situation awareness (spatial) } &VDOA &No sign. res.   & 1.63(+)  &0.81  & 1.65 &0.99 & - &- \cr
			\hline
			\multicolumn{9}{|l|}{\bf{Network-related measures}} \\
			\hline
			Localization of Router/AP  (m) & VDOA &VDOA   & 1.39(+) & 1.02 & 2.47 &1.56 & -1.83 & $<$0.1 \cr
			Situation awareness (network)  &VDOA &VDOA   & 4.5(+)  &0.97  & 3.6 &1.26 & 1.78 &$<$0.1 \cr
			Connection loss  &VDOA &VDOA  & 4/10(+) & -  & 6/10 & - & - &- \cr
			Connection quality (RSS, dBm) &VDOA &VDOA  & 2.83(+)  &1.02  & -0.54 &3.45 & 2.96 & $<$0.01 \cr
			\hline
			\multicolumn{9}{c}{}
		\end{tabular}
	\end{scriptsize}
	\label{tab:resultTable}
\end{table*}

\subsection{General results}
\label{sec:results:general}
\subsubsection{Usability} 
To measure the interface usability, we used a questionnaire with seven questions\footnote{{\revise Sample questions: I thought the interface was intuitive; I found the various functions in this interface well integrated; The interface response was slow; I thought the interface was easy to use; I enjoyed the experiment.}} each of which is scored between 1 (disagree) to 5 (agree), resulting in an overall value between 0 and 28 (5$\times$7-7) where 0 is the most difficult to use and 28 is the easiest to use. The resultant value is obtained by summing the scores of four positive questions, and subtracting the scores of three negative questions and adding an offset of 11 to obtain a positive scale of 0-28. 

A two-tailed\footnote{All the analysis made in this paper are of two-tailed nature. M indicates Mean and SD indicates Standard Deviation.} independent samples t-test was conducted to compare the usability of the VDOA and BAR. There was a significant difference in the reported usability with VDOA (M=21.3, SD=3.88) and  BAR (M=10.4, SD=2.91) conditions (t(18)=7.09, \textit{p}$<$0.01). 
This shows that the participants found the VDOA interface significantly easier to use than the BAR interface. 

\subsubsection{Mental Effort} To rate the mental effort we used the RSME scale (0 - absolutely no effort to 150 - extreme mental effort) which is essentially a one-dimensional version of the NASA-TLX scale. The resulting RSME scores of VDOA participants are M=54.9 and SD=26.25, whereas in the BAR group, the scores are M=49.1 and SD=20.2 respectively. 

It is interesting to note that there is no significant difference between the RSME scores of the two groups (t(18) = 0.55, \textit{p} = 0.58). This means that the users of the VDOA interface experienced slightly but not significantly higher mental effort than the BAR group. Thus the addition of the DOA interface did not have much impact on the cognitive load of the participants.

\subsubsection{Preference} As it is a between-subject study where each participant is assigned only to one group the evaluation of the users' preference is handled as follows. After the whole experiment and at the end of the questionnaires, we briefly explained the alternative interface (VDOA in case of BAR participants and vice versa) and asked the participant to answer the question if they would choose the alternative interface if they were given another chance. The user could answer between 1 (No) to 5 (Yes). Note the measure used is \emph{Preference to the alternative interface} and not the \emph{absolute preference to the used interface}. 
Participants that used the VDOA interface were less likely to switch to the BAR interface (i.e. not to keep using VDOA) with an average score of 1.8 (std 1.23) while significantly more participants in the BAR group preferred to switch to VDOA interface with mean score of 4.1 (std 1.45). The significance conditions are t(18) = -3.82 and \textit{p}$<$0.01.      
{\revise Note that this measure could be biased due to the general notion that humans tend to think more information is better.}

\if false
\subsubsection{Feedback} A general feedback from both the VDOA and BAR group is the suggestion to have a flashing alarm either on the screen or through an audio message when the UGV is  close to losing the signal, so that they can concentrate on the connectivity only when it's actually critical. Many users complained about the delay in the robot response, especially the turning of camera and the robot which are due to the nature of the hardware platform used.

Some participants in both groups raised concerns regarding the difficulty of multitasking to control the robot and observe the signal changes at the same time. A few participants were even surprised to see  sudden changes (or drops) in signal strengths. 

Finally, some users suggested to stabilize the color bar changes, and a few suggested some form of 3D visualization instead of 2D. A few users of the VDOA mentioned that they had difficulties in interpreting the signals from color bars due to fast changing and time-critical nature of the experiment. In fact, one user suggested to introduce the possibility of selection of different color maps (with different contrast) to help operators that suffer of severe color vision deficiency. 
\fi

\subsection{Results for the exploration scenario}
\label{sec:results:exploration}

\subsubsection{Finding symbols} Here we analyze how participants explored the maze in terms of the main exploration task which is to find as many symbols as possible.

\textit{Number of symbols found} - 
An independent samples t-test was conducted to compare the number of symbols found in the explore task. There was a significant difference in the number of objects found in VDOA (M=4.5, SD=0.97) and BAR (M=2.6, SD=2.07) conditions (t(18)=2.63, \textit{p}$<$0.05). More symbols were found with VDOA than with BAR in the actual exploration task which means the participants were able to focus on the task more productively. 

\textit{Ease of finding symbols} - 
In terms of finding symbols with ease, we asked the participants to indicate how difficult it was to find symbols during the task. The participants rated the difficulty between 1 (hard) and 5 (easy). We expected the participants that used VDOA to have found it harder to find symbols as they had to share their focus between both video and the DOA interface. However, the results suggests otherwise. The VDOA (M=4, SD=1.25) respondents reported more ease in finding symbols than the BAR ones (M=3.3, SD=1.42), but the difference is not statistically significant.

We may conclude that adding the DOA interface did not affect the operators ability to understand the spatial surroundings.
{\minor
\subsubsection{Execution time} \label{subsec_execution_time} Recall that we provided 180 seconds (3 minutes) for each participant to explore the maze. The only reason for termination before the given time limit is when the robot loses connectivity with the control station, which will be displayed in the UI as a "SIGNAL LOST" message on front of the video feed. We manually observed with a stopwatch and logged the execution times.}
{\revise As the participants did not know in advance how many symbols were there, they would normally spend all of the 180 seconds searching for symbols. The hypothesis is that VDOA users are less likely to lose connection and hence end up with longer execution times.
We found a significant difference in the execution time for VDOA (M=\unit[167.8]{s}, SD=\unit[20.53]{s}) and BAR (M=\unit[120.5]{s}, SD=\unit[67.2]{s}) conditions (t(18)= 2.13, \textit{p} = 0.047).  
See the plot on "Completion Time" in Fig.~\ref{fig:evluationresults}. 
In a typical USAR mission, being able to use the robot for searching for the
maximal amount of available time (as decided by e.g.\ time between battery changes)
 is of high importance and the VDOA interface have shown to achieve this.}

\subsubsection{Collisions} 

{\revise During the task, collisions may happen between the robot and the walls (usually when turning). This is because there was no active
collision avoidance system running, and participants may 
misinterpret the distances and sizes to obstacles in the video stream.}
We observed the number of collisions (shown in Fig.~\ref{fig:evluationresults}) with the walls of the maze in the exploration task of each participant. The average number of collisions in VDOA was 0.8 (SD 0.92) whereas in the BAR group, the mean was 0.4 (SD 0.7). Although the absolute number of collisions in VDOA was higher than the BAR group, the difference in means was not statistically significantly (t(18) = 1.09, p = 0.29) given the population size.

{\revise We believe the reason for this was twofold. First, the VDOA users ran longer missions, as they were able to stay connected longer, see Section \ref{subsec_execution_time} below. Second, they explored more difficult parts of the map, in particular the upper part where a u-turn was needed after covering the upper right corner, see Fig.   \ref{fig:coveragemap}, and as noted above most collisions occurred when turning. 

Perhaps, a better measure is the number of collisions per path length as used in \cite{de2013performance}. However, since there were many participants that had no collisions in both groups, it would not be possible to have a fair comparison with the collisions/meter metric. 
For instance, the means of the collisions per path length in the participants that had at least one collisions is VDOA (M=0.15, SD=0.01), and BAR (M=0.17, SD=0.01). On the other hand, the sum of distance traveled by all the participants that had zero collisions in VDOA is 37.7 m (4 participants) whereas it is 28.63 m in BAR (7 participants). 
}

\subsubsection{Localization of radio source} 
The participants were asked to guess the radio source {\minor(a concealed wireless router)} location and mark it on the map. We manually calculated the distance of the marked location from the actual location of the router on the map from each participant answer sheets. Following a simple rule to measure the distance, we used the Euclidean measure (shortest distance) if the marked location is in line of sight (LOS) from the router and Manhattan measure (shortest ray distance) when the marked location is in non line of sight (NLOS) from the router. From Fig.~\ref{fig:maze} we can clearly observe that the regions A and B are NLOS and all other regions are LOS. We followed this strategy not to exaggerate markings in the NLOS regions but to represent reality based on the RF propagation principles. 

The scores of localization error in each participant ranges from 0 to 6 meters. The VDOA group mean was \unit[1.39]{m} (SD \unit[1.02]{m}) and the BAR group mean was \unit[2.47]{m} (SD \unit[1.56]{m}). The difference in means are statistically significant under conditions t(18)= -1.83 and \textit{p}$<$0.1. This means that the DoA information in the UI enabled the VDOA participants to better understand the connectivity situation in real time, which is particularly helpful in increasing search and rescue mission capabilities without losing control over the robot.

\subsubsection{Coverage}
\begin{figure}[t]
	\centering
     \includegraphics[width=1.1\columnwidth, trim={4cm 0 0 0},clip]{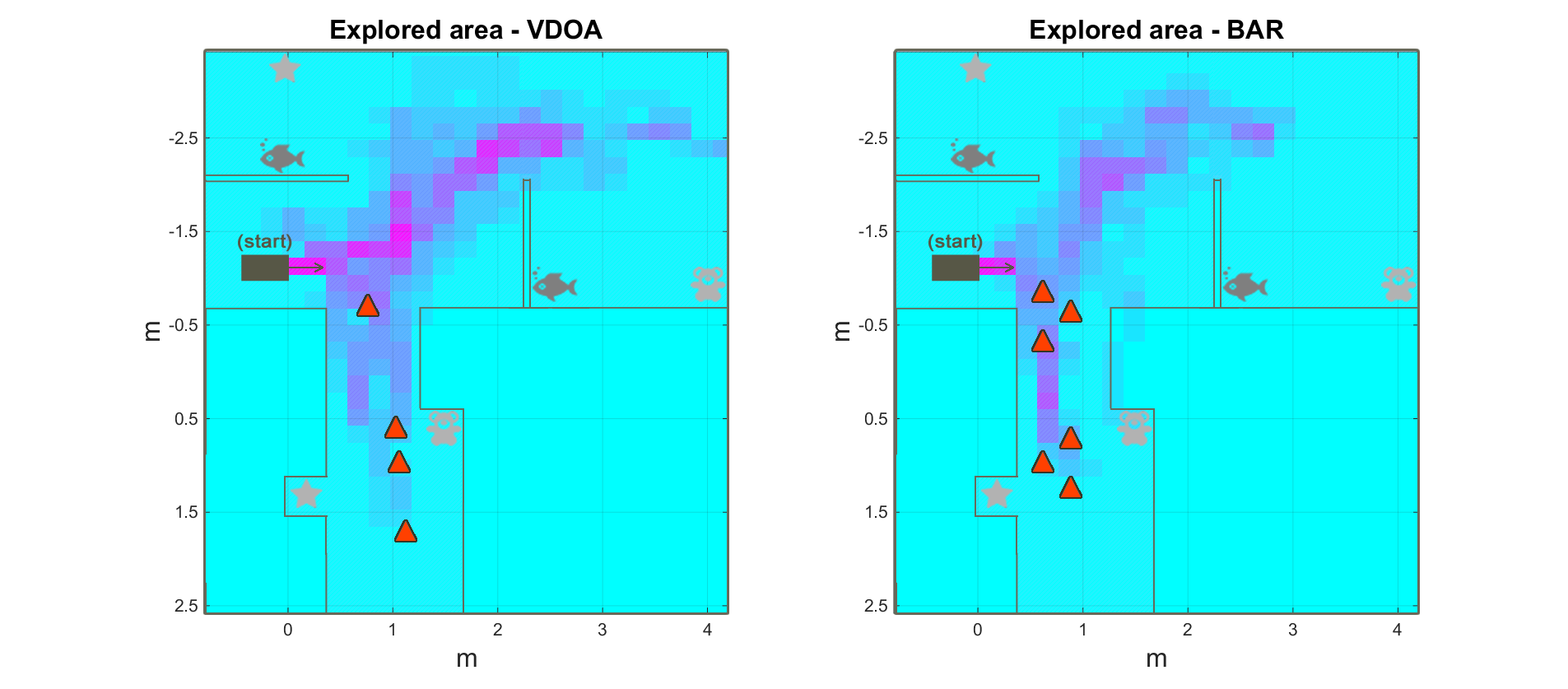}
  \caption{Two color maps showing the covered area in both groups. A lighter color represents least explored region while a darker color represents most covered region. Red triangles represent points of connection loss.
  Note how VDOA users spent more time searching the upper area, where no connection losses occured, compared to the BAR users.}
  \label{fig:coveragemap}
\end{figure}
To measure the explored areas, we discretized the maze area in 15 by 15 centimeter squares and accumulated the number of visits the robot made in each square. A graphical representation of the coverage map is shown in Fig.~\ref{fig:coveragemap} where a lighter color indicates unexplored regions. It can be seen from the map that the VDOA users spent more time exploring the regions with higher signal coverage than the BAR users. Specifically, the BAR group went more often into the low signal regions and also lost the connections much more often than the VDOA group.

We also calculated the total traveled distance by summing all the Euclidean displacements. The scores obtained are \unit[8.78]{m} (M) and \unit[2.84]{m} (SD) for VDOA and \unit[5.28]{m} (M) and \unit[3.33]{m} (SD) for BAR groups. A significant difference between the coverage area between both groups is noted (conditions: t(18) = 2.55, \textit{p}$<$0.05). Group using VDOA traveled farther in the area and covered a larger area than the BAR counterparts. 

\subsubsection{Situation awareness}
{\revise
In this study, we measured SA using a form, including both subjective (self-ratings) and objective (estimating positions in a map) components.
 In the experiments, the participants were given a fairly short time period (up to 3 minutes) to explore a fairly  small experimental area. The reason for this
is to provide a well-controlled experiment. All participants faced the same intersection, symbol placements, connectivity variations,  user interface quality, and so on, 
without having decisions regarding search strategies  influencing the data.
However, the short mission times made it difficult to
  measure SA using methods such as SAGAT \cite{endsley1988situation} (which requires questionnaire interventions during a task).  Hence we partly evaluated the SA using self-ratings from the participants, methods that,  according to \cite{gatsoulis2010}, perform equally well compared to objective methods in evaluating SA. 
}

The participants rated their confidence level in SA on a scale of 1 (lowest) to 5 (highest). There was a significant difference in how confident the participants felt that they had found all available  symbols  in the entire area using VDOA (M=4.2, SD=1.32) and BAR (M=2.5, SD=1.78) conditions (t(18)=2.42, \textit{p}=0.026). With VDOA the participants were more confident that they had  explored the entire area. Note that the participants were not informed on the number of symbols existing in the environment.

{\revise 
\textit{Spatial SA (Symbols mapping)} -
We assessed the users spatial awareness by asking the participants to mark the symbols they found during the task. Using the ground truth, we calculated the offset in the reported and the actual positions in a discrete grid map of resolution 50 cm. The offset measure was in grid spacing with 0 meaning the same grid and 9 meaning an offset of 9 grid cells. The offsets of all the found symbols were averaged to arrive at the score of each participant. A lower offset value means a better spatial awareness. Two participants in the BAR group did not find any symbols and therefore they are not considered in this analysis. We found no significant difference between the symbol mapping accuracy (spatial SA) of VDOA (M=1.63, SD=0.9) and BAR (M=1.65, SD=0.99) groups under conditions: t(16)=-0.04; \textit{p}=0.48. In Fig.~\ref{fig:evluationresults}, the presented boxplot of this measure is a normalized\footnote{We normalized the $\text{SA}_{\text{spatial}}$ score by first negating the actual score and then normalizing to the range [1,5], where higher score represents better SA.} version to correspond with the scale of other SA measures. }

On a different question we asked the participant if they felt that they had drawn the position of the radio source correctly on the map. The results (VDOA: M=4.5, SD=0.97; BAR: M=3.6, SD=1.26) reveal a significant difference between the groups (t(18)=1.78,\textit{p}$<$0.1). The VDOA group felt more aware of the network situation than the BAR group. 

Finally, we  asked participants to draw the path taken by the robot along with the final orientation after they finished the task and asked a question how confident they felt in marking the path. Although the VDOA group (M=4.5, SD=0.85) had higher confidence than the BAR group (M=4, SD=1.25) in general, there was no significant difference. This may attribute to the fact that both group used the same FLC control and may mean that having an additional indicator for directional wireless connectivity does not inhibit operator awareness of the robots position and orientation. 

\subsection{Network parameters}
\label{sec:results:network}

\subsubsection{Connection loss}
The connection loss measure is directly related to the execution time, as the exploration task was terminated before the timeout only when the participant lost the connection. Therefore one might expect that the analysis of the execution time holds also for the connection loss measure. However, a t-test on how many participants lost connection {\minor (4 out of 10 for VDOA and 6 out of 10 for BAR)} during the study showed no significant difference between the means of VDOA (M=0.4, SD=0.52) and BAR (M=0.6, SD=0.52). 
One reason for this might be that when mission time grows to infinity, the chances of losing connectivity at some point 
tends to one, regardless of what interface is used. {\minor Furthermore, after detecting a certain amount of symbols, VDOA users tended to adopt a riskier strategy, pushing the robot to explore the edges of the poorly connected area and causing a connection loss (4 out of 10), see Fig.~\ref{fig:coveragemap}.}

\subsubsection{Overall connection quality}
\begin{figure}[t]
  \center
     \includegraphics[width=1.1\columnwidth, trim={5cm 0 0 0},clip]{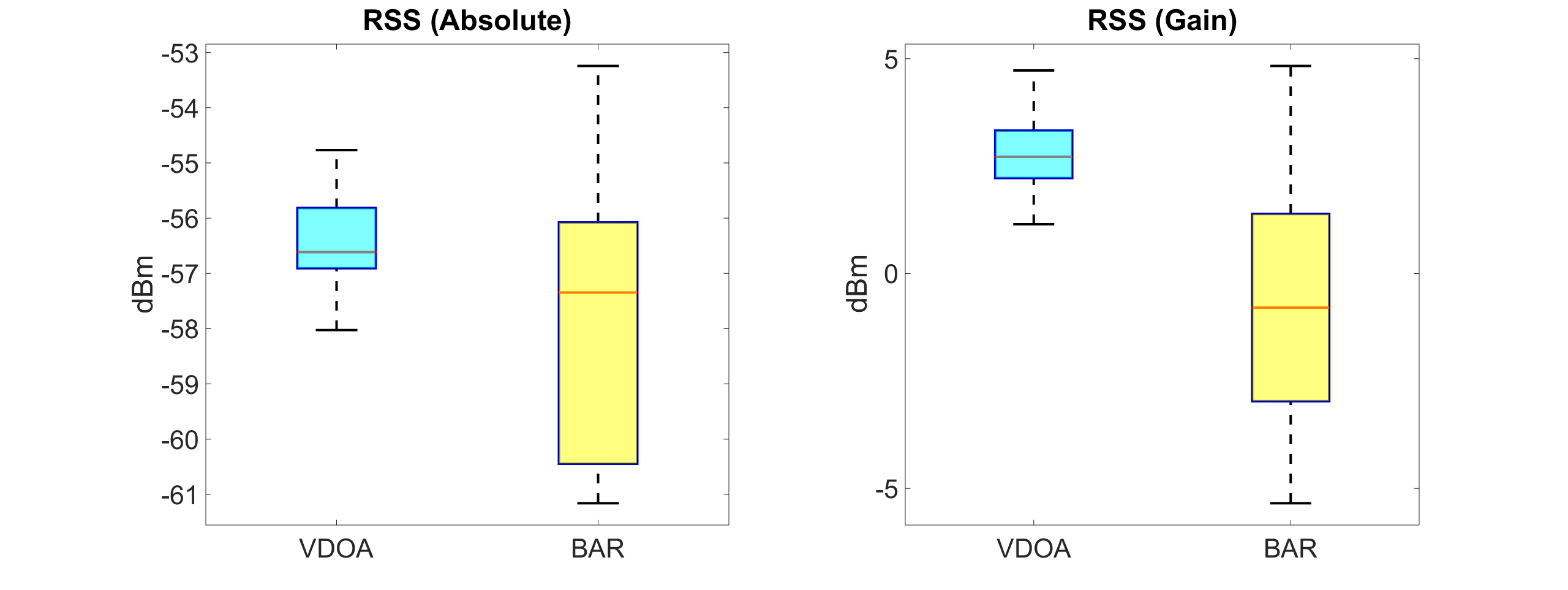}
  \caption{{Boxplots of the average RSS values and the average RSS gains in both groups.}}
  \label{fig:RSSplot}
\end{figure}
We used the RSS from the wireless adapter as a measure of overall connection quality. As we are interested in the improvement in connection quality from the starting position, we calculate the difference in the RSS from the initial RSS values and calculated the RSS gain averaged over the entire duration of the exploration task by each participant. In this way we mitigate the influences of temporal variations and effects of influences due to network traffic conditions during the day. Positive values indicate that there is an improvement in the RSS values and negative values indicates the opposite.

We found significant difference in RSS gain between VDOA (M=\unit[2.83]{dBm}, SD=\unit[1.02]{dBm}) and BAR (M=\unit[-0.54]{dBm}, SD=\unit[3.45]{dBm}) under conditions t(18)=2.96 and \textit{p}$<$0.01. 
These values are for the RSS of the central receiver which is  used to transfer data to and from the robot. The results are  the same regardless of which receiver we consider including the mean of all the receiver RSS values. 

Fig.~\ref{fig:RSSplot} shows the boxplots of both absolute RSS values and the RSS gains for both groups. It can be seen that in general the VDOA group maintained higher RSS than the BAR group.
{\revise Recall from Sec.~\ref{sec:task} that the users had three options at the beginning of the task: go straight (high connectivity region - D, F, G, H), turn left (medium connectivity region - C and E), and turn right (poor connectivity region - A and B). Two symbols were placed in each of these regions, as can be seen in Fig.~\ref{fig:coveragemap}. The VDOA users mostly preferred going straight and left than the BAR users due to the additional 2D connectivity information. However, after exploring those regions they proceeded to explore the poor connectivity region with caution.}

\section{Discussions}
\label{sec:discussions}

In this section we first discuss the results in relation to the claims we made in Section~\ref{sec:evaluation}. 
The more upsides and downsides we can confirm, the stronger support we have for the 
corresponding claim. We then discuss the results in more detail and finally make some general remarks.

Regarding Claim 1, \emph{`VDOA leads to UGV trajectories in  higher signal strength regions'}, the upsides U11, U12 and U13 were confirmed. As there is no significant difference in the number of collisions and the ease of finding symbols, we can not confirm the downside D11. These results support Claim 1.

Claim 2, \emph{`VDOA provides better situation awareness'}, is more complex.
U21 is partially confirmed, with participants being equally aware of the path traversed, but VDOA users being more confident to have explored the area.
U22 is confirmed, with more symbols found by the VDOA users.
U23 is partially confirmed, with the same accuracy of mapping found symbols, but better accuracy of radio source localization.
D21 is not confirmed.
To conclude, the results partially support Claim~2.

In Claim 3, \emph{`VDOA improves user experience'}, all the upsides U31, U32, U33 were confirmed. This strongly supports  Claim 3.

Regarding Claim 4, \emph{`BAR increases focus and concentration on the actual task'}, we are unable to confirm the upsides U41 and U42 because we did not find significant difference in the measures. Also the upside U43 was refuted because the BAR was neither a preferred system nor rated higher in usability. On the other hand, we can confirm the downsides D41 and D42. Consequently we only have a weak support for Claim 4.

Looking at the support for all claims, and in particular the fact that
all upsides and no downsides of VDOA (Claims 1-3) were confirmed, and all downsides but no upsides of BAR (Claim 4) were confirmed,
we can conclude that VDOA is preferable to BAR
in wireless teleoperation of UGVs.

Regarding the general results, in Section~\ref{sec:results:general},
we note that VDOA was considered easier to use, similar in terms of mental effort required and preferred by a majority of the operators.
We believe that these advantages are due to the fact that 
 the DoA information is added in the periphery of the video feed in a way that  can be accurately and easily processed.

Regarding the exploration results, in Section~\ref{sec:results:exploration},
we note that VDOA resulted in more symbols found, a longer travelled distance and time, improved accuracy in locating the radio source, similar accuracy in marking found symbols on a map, and a slight increase in number of collisions.
We believe that these advantages are due to the fact that DoA information is 
very important when making decisions close to the connectivity threshold.
Manually estimating the DoA using the information provided in the BAR interface is possible,
but probably associated with a significant cost in terms of mental load and mission time,
and impossible to do with an accuracy similar to the one observed in Section \ref{sec:Experiments} ($<12$ degrees).
The reason for users of the BAR interface losing connection with the UGV was probably that
they were not able to manually estimate the DoA accurately enough. Without a reliable estimate, a natural reaction when running into a low RSS area is to move back to the area just visited, but that strategy has a negative impact on the exploration objective.

Regarding the network results, in Section~\ref{sec:results:general}, VDOA resulted in a higher overall connection quality. As noted above, having access to a DoA estimate enables the operator to choose paths that takes both the connectivity and exploration objectives into account.
{\revise Thus, the DoA information is not guiding the robot, instead it is enabling robot operations in low connectivity regions and in the regions close to the connectivity threshold. 
The operator chooses where to go to perform the search. With the VDOA information, the operator can predict the risk better. If entering a room presents a high risk of losing connectivity, the operator can still enter if the potential information gain is worth it. With BAR, the operator might lose connectivity without understanding that the risk was there, as shown in the user study, see Fig.~\ref{fig:coveragemap}}.
In USAR missions, where staying connected with the robot is critical for saving lives, this VDOA interface could play a vital role.}

{\revise 
Finally, from a scientific point of view, we would like to note that this study provides a slight elaboration {\minor of the identifying, measuring and analyzing SA variables relevant to the context}. As shown in this application, there can be an interaction of dynamic environmental conditions (e.g., network coverage) and robot capabilities (e.g., tele-operation) that affect task performance. So, SA support should not only focus on the perception, comprehension and prediction of events and states that directly relate to the primary  task (e.g., obstacles when navigating), but also focus on the availability and dependencies of the required resources for the task execution.
Furthermore, the peripheral color bar in the display provides a general UI pattern for the corresponding SA-support, hardly interfering with the primary task,
and easily extendable to other forms of scalar field measurements, such as temperature, gas density or sound volume.
}

\section{Conclusions}
\label{sec:conclusions}

In this paper, we proposed a way of estimating DoA of the radio signal and a way of including this information in a UGV teleoperation interface.
We also investigated the quality of the estimates and conducted a user study showing that the new interface resulted in improved performance in an exploration scenario.

In the technical tests, we showed that the DoA estimates had a mean error of less than 12 degrees, and were useful for predicting changes in RSS values over a typical mission trajectory.

In the user study,
the benefits of the new interface, that incorporates directional wireless connectivity information in the Free Look Control interface, were compared to the standard "signal bar" representation of the wireless signal level used in many modern UGV user interfaces for remote teleoperation. 

We conducted a between-subjects user evaluation with 24 participants and were able to  analyze 20 of them with 10 in each group (VDOA and BAR) and found that the new interface (VDOA) {\revise partially improves users' situation awareness and significantly reduces} connection loss with the robot. This is especially useful in robot aided USAR situations 
where connection loss has a huge impact on  mission performance.

A possible extension of this research is to integrate the proposed interface in an augmented reality display system \cite{kruckel2015} to represent the wireless connectivity in a 3D fashion as some participants suggested in their  feedback. Additionally, the directional antennas used in this study can also be exploited for communication redundancy, offering advantages such as increased coverage, stable connections, and coverage in elevated regions \cite{Yasushi2011}.
Finally, we believe that the VDOA interface can be easily adapted to both teleoperated maritime and unmaaerial vehicles.

\section*{Acknowledgments}
The authors gratefully acknowledge funding under the European Union's seventh framework program (FP7), under grant agreements FP7-ICT-609763 TRADR.

\bibliography{tradrRefs,ugvReferenser,WirelessRef}
\bibliographystyle{IEEEtran}

\end{document}